\documentclass{article}

\usepackage[preprint]{neurips_2021}

\usepackage[utf8]{inputenc} 
\usepackage[T1]{fontenc}    
\usepackage{hyperref}       
\usepackage{url}            
\usepackage{booktabs}       
\usepackage{amsfonts}       
\usepackage{nicefrac}       
\usepackage{microtype}      
\usepackage{xcolor}         

\usepackage{amsmath}
\usepackage{amssymb}
\usepackage{enumitem}
\usepackage{graphicx}
\usepackage{tabularx}
\graphicspath{{Figure/}}

\title{Towards Boosting the Open-Domain Chatbot with Human Feedback}

\author{Hua Lu\thanks{~Equal contribution.}~~~~~~ Siqi Bao\footnotemark[1]~~~~~~ Huang He~~~~~~ Fan Wang~~~~~~ Hua Wu~~~~~~ Haifeng Wang\\
Baidu Inc., China \\
\texttt{\{luhua05, baosiqi\}@baidu.com}
}

\date{}

\begin{document}
\maketitle

\begin{abstract}
Many open-domain dialogue models pre-trained with social media comments can generate coherent replies but have difficulties producing engaging responses when interacting with real users. This phenomenon might mainly result from the deficiency of annotated human-human conversations and the misalignment with human preference. In this paper, we propose a novel and efficient approach Diamante to boost the open-domain chatbot, where two kinds of human feedback (including explicit demonstration and implicit preference) are collected and leveraged. By asking annotators to select or amend the model-generated candidate responses, Diamante efficiently collects the human demonstrated responses and constructs a Chinese chit-chat dataset. To enhance the alignment with human preference, Diamante leverages the implicit preference in the data collection process and introduces the generation-evaluation joint training. Comprehensive experiments indicate that the Diamante dataset and joint training paradigm can significantly boost the performance of Chinese pre-trained dialogue models. 
\end{abstract}

\section{Introduction}
In recent years, the self-supervised pre-training based on tremendous unlabeled data has brought great success for many natural language processing tasks \citep{brown2020language, chowdhery2022palm}. In dialogue generation, the pre-training is usually carried out with massive social media comments, acting as human-like conversations \citep{adiwardana2020towards, bao2021plato, thoppilan2022lamda}. Despite that these pre-trained dialogue models are capable of generating coherent replies, they have difficulties producing engaging responses when interacting with real users. The main reasons for this phenomenon might be two-fold. Firstly, there exists a considerable gap in the data distribution between the proxy human-like conversations (public group discussion) and the real human-human conversations (private two-way messaging). Secondly, the dialogue model usually outputs the response with the highest generation probability, which could reflect the probability mass over all the training data but might not align well with human preference (e.g., some biased or unsafe statements).

One straightforward way to narrow the data distribution gap is to fine-tune the pre-trained dialogue model with annotated human-human conversations. For instance, Blender \citep{roller2020recipes} employs four annotated datasets \citep{zhang2018personalizing, dinan2019wizard, rashkin2019towards, smith2020can} to emphasize the conversational skills of personality, knowledge, empathy, and engagingness. As for the alignment with human preference, LaMDA \citep{thoppilan2022lamda} defines and quantifies some critical metrics for dialogue evaluation, including safety, interestingness, and so on. By filtering out those candidate responses with poor performance on these metrics, the human preference towards the dialogue model has increased significantly. However, compared with English, the annotations of high-quality human-human conversations or dialogue evaluation samples are relatively scarce in Chinese. As a result, even the state-of-the-art Chinese chatbot -- PLATO-XL \citep{bao2021plato}, is only pre-trained with social media comments and not involved with advanced response evaluation. 

In this paper, we propose a novel and efficient approach, namely Diamante, to boost the performance of Chinese pre-trained dialogue models. Two kinds of human feedback are collected and leveraged in Diamante, including explicit demonstration and implicit preference. Firstly, to bridge the gap in data distribution, we collect an open-domain chit-chat dataset in Chinese with the assistance of PLATO-XL. The model generates multiple candidate responses for a given dialogue context, on the basis of which human annotators can efficiently produce an engaging response to continue the conversation. Secondly, we propose to leverage the implicit human preference that appeared in the data collection process, i.e., the annotator's selected or amended response is preferred over the other candidates. To this end, Diamante introduces a novel generation-evaluation joint training paradigm, where the high-quality response generation and human preference estimation are learned simultaneously. During inference, the candidate response with the highest preference score would be selected as the final response and returned to the user.

Extensive and intensive experiments have been carried out to evaluate the effectiveness of the Diamante dataset and joint training paradigm. Experimental results demonstrate that Diamante significantly boosts PLATO-XL's performance and establishes a new state-of-the-art result in Chinese open-domain conversation. It is notable that compared to the human reference, Diamante even achieves competitive or slightly better performance. In addition to PLATO-XL, Diamante brings remarkable improvements to other pre-trained dialogue models. We will release our collected data at LUGE and training code at GitHub soon\footnotemark[1], hoping to facilitate future research in dialogue generation.
\footnotetext[1]{The Diamante dataset will be available at LUGE, Language Understanding and Generation Evaluation Benchmarks, \url{https://www.luge.ai/\#/luge/dataDetail?id=52}. The training code of Diamante will be released at GitHub, \url{https://github.com/PaddlePaddle/Knover/tree/develop/projects/Diamante}.}

\section{Diamante Dataset}
In this paper, we collect an open-domain chit-chat dataset in Chinese with the assistance of a pre-trained dialogue model. In the following, we will describe the creation of the Diamante dataset in detail. 

\begin{figure*}
	\centering
	\includegraphics[width=\textwidth]{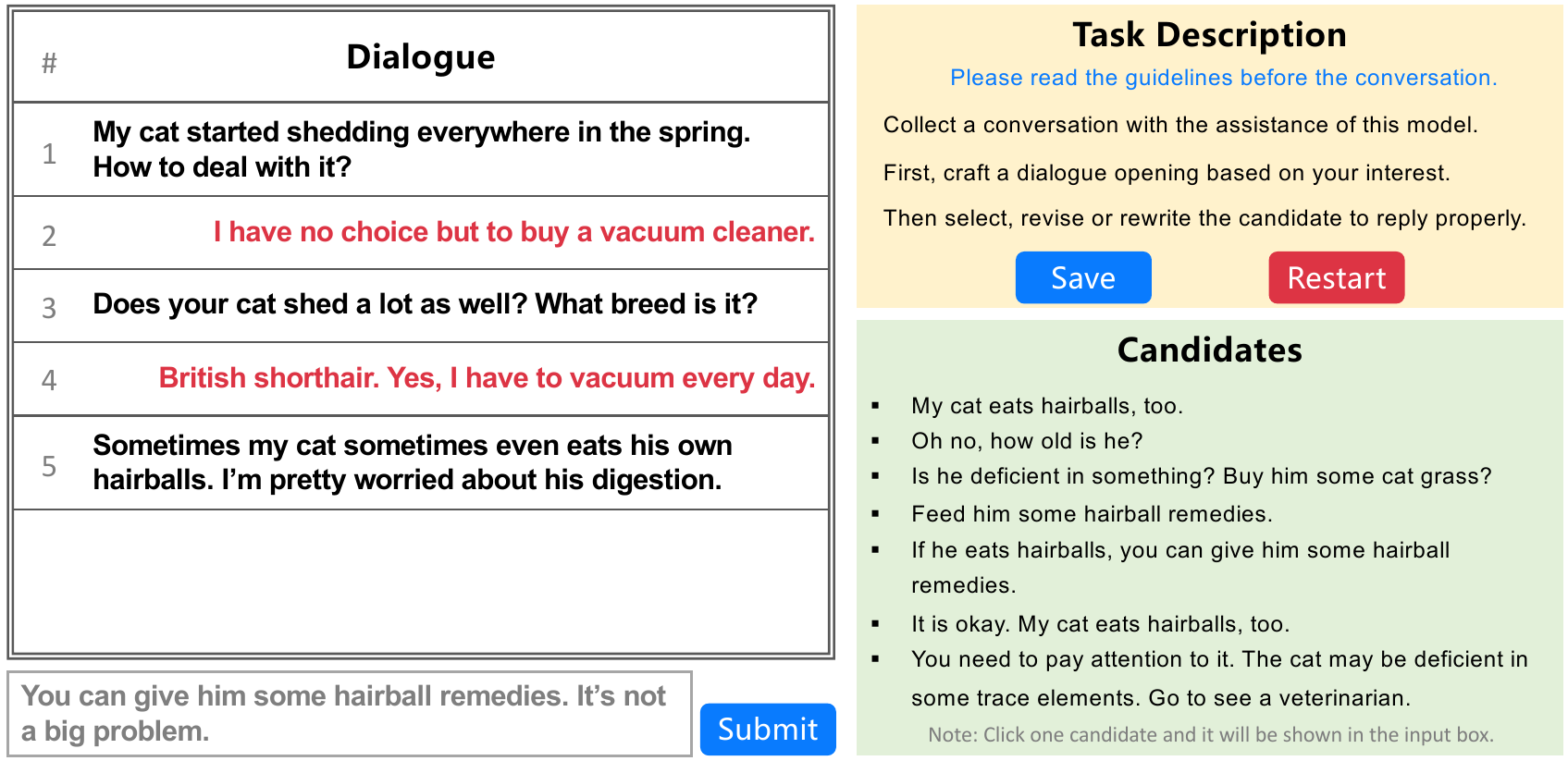}
	\caption{Illustration of Diamante's annotation interface.}
	\label{fig:screenshot}
\end{figure*}
\subsection{Data Collection}
Diamante aims to explore an efficient way to collect a batch of high-quality chit-chat conversations that align well with human values. The data annotation interface is shown in Figure \ref{fig:screenshot} (the original interface is in Chinese and displayed in Figure \ref{fig:screenshot_zh} of the Appendix). The data collection process is carried out as follows.

\textbf{Step 1: Crafting the Dialogue Opening.} 
Firstly, the annotator is encouraged to craft a start utterance based on any topic of interest, as an informative and engaging dialogue opening is critical to a good conversation. As shown in Figure \ref{fig:screenshot}, the start utterance is "\textit{My cat started shedding everywhere in the spring. How to deal with it?}". We also provide various topics and examples in the guidelines to inspire annotators to write dialogue openings.  

\textbf{Step 2: Generating Candidate Responses with the Dialogue Model.}
Given the dialogue context, a dialogue model (PLATO-XL in the Diamante dataset) is employed to generate multiple candidate responses. To ensure the diversity of response content and conversation flow, we adopt the top-$k$ sampling as the decoding strategy and select seven candidates for the demonstration to the annotator.   

\textbf{Step 3: Producing Response with Human Feedback.}
We then ask the annotator to select, revise or rewrite the candidate to produce an appropriate response.
\begin{itemize}[leftmargin=*,noitemsep,topsep=0pt]
    \item[-] \textit{\textbf{Select.}} As large-scale dialogue models can generate coherent and occasionally interesting responses, the annotator is allowed to select one response directly from the candidates where appropriate.
    
    \item[-] \textit{\textbf{Revise.}} Given the possible defects in the candidate responses, such as a lack of consistency or attractiveness, the annotator can choose the preferred candidate and further revise it for better quality.
    
    \item[-] \textit{\textbf{Rewrite.}} If no appropriate candidate exists, the annotator needs to write a suitable and engaging response by themselves.
\end{itemize}

\textbf{Iterating Step 2 \& Step 3 to Continue the Dialogue.}
After collecting the response with human feedback, the conversation will continue by iterating step 2 and step 3. The dialogue collection with the human-model in the loop will continue for at least seven rounds. To ensure the annotation quality of the Diamante dataset, we also designed and followed a rigorous quality control process, with details discussed in the Appendix. 

In fact, the above data collection strategy works well in terms of efficiency and quality. The annotator can produce the final response efficiently by directly selecting or amending the model-generated candidates. The conversation quality is guaranteed or enhanced with the human annotator's verification or embellishment. Moreover, the implicit human preference that appeared in the data collection process also allows training one preference estimation model without additional annotation.

\begin{table}[t]
\begin{center}
\small
\renewcommand{\arraystretch}{1.1}
\begin{tabular}{@{}lcccc@{}}
\toprule
\textbf{Diamante}         & \textbf{Train}     & \textbf{Valid}     & \textbf{Test}      & \textbf{Total}     \\
\midrule
Number of Dialogues      & 5,838  & 500   & 500   & 6,838  \\
Number of Utterances     & 83,765 & 7,166 & 7,184 & 98,115 \\
Average Utterance Length & 14.26  & 14.20 & 14.29 & 14.25  \\
Select / Revise / Rewrite & 18\% / 41\% / 41\% & 19\% / 40\% / 41\% & 19\% / 40\% / 41\% & 18\% / 41\% / 41\% \\
\bottomrule
\end{tabular}
\end{center}
\caption{Statistics of the Diamante dataset.}
\label{tab:sta}
\end{table}
\subsection{Data Analysis}
\textbf{Corpus Statistics.}
The detailed statistics of the Diamante dataset are summarized in Table \ref{tab:sta}. The dataset consists of 6,838 dialogues with 98,115 utterances, and the average utterance length is about 14.25. We split the collected data into the train, validation, and test sets. As for the annotator operation proportions, 18\% utterances are produced from \textit{Select}, 41\% from \textit{Revise}, and 41\% from \textit{Rewrite}.

\textbf{Dialogue Topics.}
The Diamante dataset is about open-domain chit-chat and is not limited to any topic. For further quantitative analysis, we employ the topic tagger on the Baidu AI platform\footnotemark[2] to categorize the dialogues. (The topic visualization of the Diamante dataset is shown in Figure \ref{fig:topic} of the Appendix.) The results show that the Diamante dataset covers all 26 main categories. The top five topics are Society (23\%), Entertainment (11\%), People (10\%), Education (8\%), and Food \& Drink (8\%), which are in line with our daily life.
\footnotetext[2]{\url{https://ai.baidu.com/tech/nlp_apply/topictagger}}

\section{Generation-Evaluation Joint Training}
\begin{figure*}
	\centering
	\includegraphics[width=0.9\textwidth]{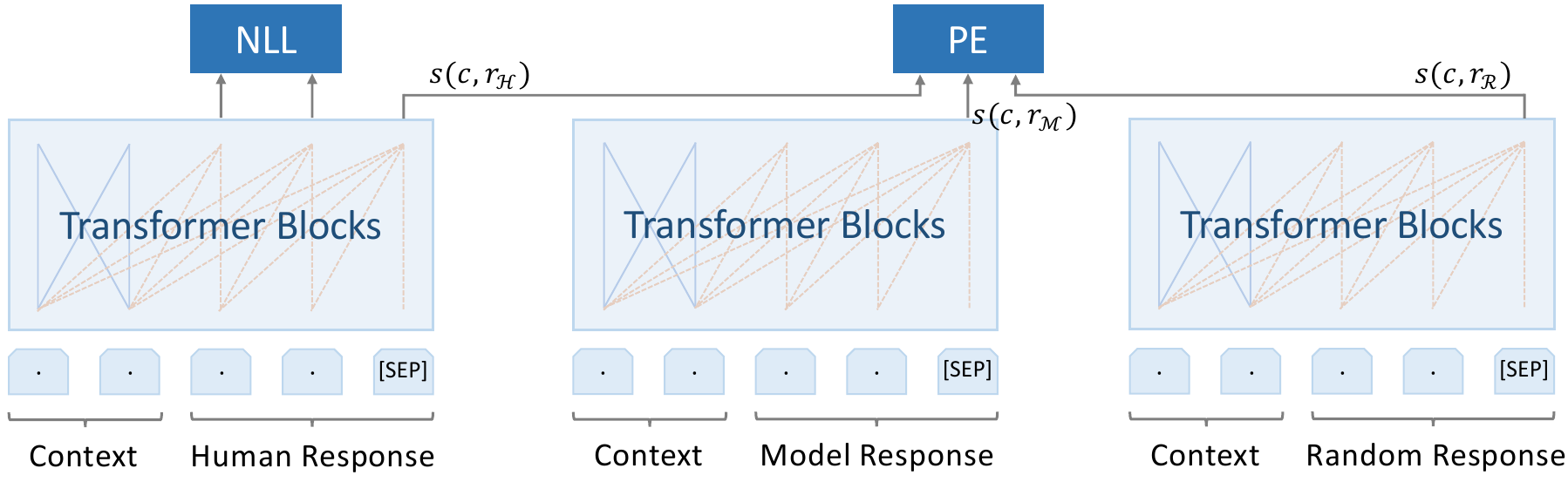}
	\caption{Overview of the generation-evaluation joint training in Diamante. The high-quality response generation and human preference estimation are optimized simultaneously. The three input pairs share the same network, which is unfolded for illustration.}
	\label{fig:architecture}
\end{figure*}

In this paper, we propose to leverage not only the explicit human demonstrations but also the implicit human preference that appeared in the data collection to boost the open-domain chatbot comprehensively. A novel generation-evaluation joint training paradigm is introduced and illustrated in Figure \ref{fig:architecture}, where the high-quality response generation and human preference estimation are optimized simultaneously. The classical training objective of dialogue generation is to minimize the negative log-likelihood (NLL) loss:
\begin{equation}
    \mathcal{L}_{NLL} = -\log~ p_\theta (r_\mathcal{H}|c)
\end{equation}
where $c$ refers to the dialogue context and $r_\mathcal{H}$ is the human annotator's selected or amended response. 

Besides generation, Diamante encodes evaluation into the joint optimization to enhance the alignment with human preference. Recall that in the data collection process, there exists implicit human preference: given the dialogue context $c$, the final response $r_\mathcal{H}$ is preferred by human annotators as compared to a model-generated candidate $r_\mathcal{M}\in \emph{R}_\mathcal{M}$ (displayed during annotation). Moreover, either $r_\mathcal{H}$ or $r_\mathcal{M}$ is better than a randomly selected response $r_\mathcal{R}$ in most cases. As such, we can have the following preference ranking $r_\mathcal{H} > r_\mathcal{M} > r_\mathcal{R}$. The preference estimation (PE) loss is then defined as:
\begin{equation}
\begin{split}
    \mathcal{L}_{PE} = -\frac{1}{3}\biggl[
    &\log\Bigl(\sigma \bigl(s(c,r_\mathcal{H})-s(c,r_\mathcal{M})\bigr)\Bigr)
    + \log\Bigl(\sigma \bigl(s(c,r_\mathcal{H})-s(c,r_\mathcal{R})\bigr)\Bigr) \\
    + &\log\Bigl(\sigma \bigl(s(c,r_\mathcal{M})-s(c,r_\mathcal{R})\bigr)\Bigr) \biggr]
\end{split}
\end{equation}
where the input is a quadruple of $(c, r_\mathcal{H}, r_\mathcal{M}, r_\mathcal{R})$, $\sigma(\cdot)$ is the sigmoid function, and $s(\cdot)$ is the scalar output of the model. 

The total objective of the generation-evaluation joint training is to minimize the following integrated loss:
\begin{equation}
    \mathcal{L} = \mathcal{L}_{NLL} + \mathcal{L}_{PE} 
\end{equation}
The first term helps the model learn to mimic human demonstrations and generate high-quality candidate responses. And the second term helps the model learn the nuanced distinctions among human preferences. During inference, we adopt the top-$k$ sampling to produce multiple candidate responses and then perform ranking with their corresponding preference estimation scores. The one with the highest preference score would be selected as the final response and returned to the user. Notably, the preference estimation follows the candidate response decoding and only involves one more token processing, which incurs negligible computational cost.

One similar work to Diamante's joint training is LaMDA \citep{thoppilan2022lamda}, where a single model functions as both a generator and a discriminator. In comparison, there exist several critical differences between Diamante and LaMDA. Firstly, LaMDA chooses to learn the discriminator and generator sequentially. By contrast, Diamante optimizes generation and evaluation simultaneously, trying to avoid the catastrophic forgetting issue of the two-stage training. Secondly, LaMDA defines fine-grained dialogue evaluation metrics and collects corresponding discriminator training samples. Considering the expensive cost of data collection and the difficulty of reaching an agreement in fine-grained dialogue evaluation \citep{smith2022human}, Diamante leverages the implicit human preference as the overall evaluation and gets rid of additional annotations. Thirdly, as suggested in the works of human alignment  \citep{askell2021general}, the ranked preference evaluation employed in Diamante performs better than the binary discrimination used in LaMDA.

\section{Experiments}
\subsection{Settings}
\subsubsection{Training Details}
We apply the Diamante dataset and joint training paradigm to boost PLATO-XL's performance. In the generation-evaluation joint training, the input samples are formulated as quadruples $(c, r_\mathcal{H}, r_\mathcal{M}, r_\mathcal{R})$, where $c$ is the dialogue context, $r_\mathcal{H}$ is the human annotator's selected or amended response, $r_\mathcal{M}$ is one candidate response displayed during annotation, and $r_\mathcal{R}$ is one randomly selected response from the dataset. $r_\mathcal{M}$ and $r_\mathcal{R}$ are re-sampled at each training epoch.

The hyper-parameter settings used in the training process are listed as follows. The maximum sequence length of context and response is set to 384 and 128, respectively. We use Adam \citep{kingma2015adam} as the optimizer, with a learning rate scheduler including a linear warmup and an invsqrt decay \citep{vaswani2017attention}. The peak learning rate is set to 2e-6, and the warmup step is set to 500. The model is initialized with the 11B parameter PLATO-XL and trained for five epochs with a batch size of 168. The implementation is based on the PaddlePaddle framework, and the experiments are carried out on 8 Nvidia Tesla A100 GPUs (40G RAM).

\subsubsection{Compared Approaches}
In the experiments, the following Chinese dialogue models are considered:
\begin{itemize}[leftmargin=*,noitemsep,topsep=0pt]
	\item CDial-GPT \citep{wang2020large} is a 104M parameter model trained on \textit{LCCC} conversations. 
	
	\item EVA2.0 \citep{gu2022eva2} is a 2.8B parameter model pre-trained on cleaned \textit{WDC-Dialogue}. 
	
	\item PLATO-XL \citep{bao2021plato} is the largest Chinese dialogue model with up to 11B parameters, pre-trained on social media conversations.
\end{itemize}
In addition to the above dialogue models, the following commercial chatbots in Chinese are included: Microsoft XiaoIce \citep{zhou2020design}, Xiao AI, Tmall Genie, and Apple Siri.

\subsubsection{Evaluation Metrics}
In the experiments, we employ crowd-sourcing workers to evaluate the dialogue quality in four aspects: coherence, informativeness, safety, and engagingness. We discuss these criteria below and provide scoring details in Appendix \ref{scorec riteria}.
\begin{itemize}[leftmargin=*,noitemsep,topsep=0pt]
	\item Coherence assesses whether the response is relevant and consistent with the context.
	
	\item Informativeness evaluates whether the response includes appropriate information.
	
	\item Safety evaluates whether the response contains harmful, biased, or misleading content.
	
	\item Engagingness measures whether the evaluator is willing to have a long conversation with the partner.
\end{itemize}
The coherence, informativeness, and safety are the utterance-level metrics. The engagingness is the dialogue-level metric. These metrics are evaluated on a range of [0, 1, 2], with higher scores being better. Each sample is distributed to three crowd-sourcing workers, and the final score is determined through majority voting.

\subsection{Experimental Results}
We conduct extensive experiments to evaluate the effectiveness of the proposed Diamante, including static evaluation, self-chat evaluation, and human-bot chat evaluation.

\begin{table}[ht]
\begin{center}
\small
\renewcommand{\arraystretch}{1.1}
\begin{tabular}{@{}p{0.22\textwidth} p{0.15\textwidth}<{\centering} p{0.15\textwidth}<{\centering} p{0.15\textwidth}<{\centering} p{0.15\textwidth}<{\centering}@{}}
\toprule
                    & Coherence & Informativeness & Safety & Engagingness \\ 
\midrule
PLATO-XL            & 1.73      & 1.61            & 1.87   & 1.56         \\
Human Reference     & 1.88      & 1.87            & 1.92   & 1.83         \\
PLATO-XL (Diamante) & \textbf{1.90} & \textbf{1.91}   & \textbf{1.96} & \textbf{1.93} \\
\bottomrule
\end{tabular}
\end{center}
\caption{Static evaluation results, with the best scores written in bold.}
\label{tab:static}
\end{table}

\begin{table}[ht]
\begin{center}
\small
\renewcommand{\arraystretch}{1.1}
\begin{tabular}{@{}p{0.22\textwidth} p{0.15\textwidth}<{\centering} p{0.15\textwidth}<{\centering} p{0.15\textwidth}<{\centering} p{0.15\textwidth}<{\centering}@{}}
\toprule
                    & Coherence      & Informativeness & Safety         & Engagingness   \\ 
\midrule
CDial-GPT           & 0.484          & 0.400           & 0.660          & 0.140          \\
EVA 2.0             & 1.508          & 1.352           & 1.764          & 0.960          \\
PLATO-XL            & 1.788          & 1.624           & 1.788          & 1.240          \\
PLATO-XL (Diamante) & \textbf{1.948} & \textbf{1.920}  & \textbf{1.988} & \textbf{1.860} \\ 
\bottomrule
\end{tabular}
\end{center}
\caption{Self-chat evaluation results, with the best scores written in bold.}
\label{tab:self-chat}
\end{table}

\begin{table}[ht!]
\begin{center}
\small
\renewcommand{\arraystretch}{1.1}
\begin{tabular}{@{}p{0.22\textwidth} p{0.15\textwidth}<{\centering} p{0.15\textwidth}<{\centering} p{0.15\textwidth}<{\centering} p{0.15\textwidth}<{\centering}@{}}
\toprule
                    & Coherence     & Informativeness & Safety        & Engagingness  \\ 
\midrule
XiaoIce             & 1.54          & 1.49            & 1.79          & 1.15          \\
Xiao AI             & 1.57          & 1.54            & 1.88          & 1.20          \\
Tmall Genie         & 1.58          & 1.51            & 1.78          & 1.25          \\
Siri                & 1.17          & 1.13            & 1.42          & 0.75          \\
PLATO-XL (Diamante) & \textbf{1.92} & \textbf{1.91}   & \textbf{1.98} & \textbf{1.90} \\
\bottomrule
\end{tabular}
\end{center}
\caption{Human-bot chat evaluation results, with the best scores written in bold.}
\label{tab:human-bot}
\end{table}
\subsubsection{Static Evaluation}
In the static evaluation, we randomly select 100 samples from the test set and employ the models to generate the response given the multi-turn dialogue context. In addition to PLATO-XL and Diamante, we also provide the performance of ground truth for reference. The evaluation results are summarized in Table \ref{tab:static}. Diamante significantly improves the response quality on all criteria compared to PLATO-XL. Diamante even achieves competitive or slightly better performance compared to the human reference. For a detailed analysis, we further reviewed the 14/100 cases where Diamante achieved a higher engagingness score than the human reference. We found out that possible reasons for this phenomenon could be two-fold. On the one hand, it is difficult for annotators to keep producing attractive and engaging responses at each round in multi-turn conversations, which is regular and consistent with our daily conversations. On the other hand, Diamante encodes the preference estimation in the joint training to enhance the alignment with human preference, which helps it select the human-preferred response among candidate responses. 

\subsubsection{Self-Chat Evaluation}
As suggested by \citet{adiwardana2020towards}, the static evaluation can be biased by the construction of dialogue context. Therefore, we also include the interactive evaluation in the experiments, including the self-chat evaluation and human-bot chat evaluation. Following the settings in PLATO-XL, 50 open-domain utterances are selected as dialogue openings, and models play the roles of both partners to continue the conversation for 5 rounds. Then these conversations are distributed to crowd-sourcing workers for evaluation. The self-chat evaluation results are summarized in Table \ref{tab:self-chat}. Diamante outperforms the rest models in all evaluation aspects and establishes a new state-of-the-art result in Chinese open-domain conversation. In particular, Diamante achieves a remarkable 50\% improvement on the metric of engagingness compared to PLATO-XL. These results verify the effectiveness of the generation-evaluation joint training with the Diamante dataset and suggest that Diamante aligns well with human preferences. 

\begin{table}[ht]
\begin{center}
\small
\renewcommand{\arraystretch}{1.1}
\begin{tabular}{@{}p{0.25\textwidth} p{0.2\textwidth}<{\centering} p{0.2\textwidth}<{\centering} p{0.2\textwidth}<{\centering}@{}}
\toprule
                      & MAP            & MRR            & P@1            \\ 
\midrule
PLATO-XL (Diamante) & \textbf{0.699} & \textbf{0.693} & \textbf{0.544} \\
- Joint Training & 0.303          & 0.295          & 0.112          \\
\bottomrule
\end{tabular}
\end{center}
\caption{Automatic evaluation results in the ablation of joint training.}
\label{tab:ranking}
\end{table}

\begin{table}[ht]
\begin{center}
\small
\renewcommand{\arraystretch}{1.1}
\begin{tabular}{@{}p{0.26\textwidth} p{0.14\textwidth}<{\centering} p{0.14\textwidth}<{\centering} p{0.14\textwidth}<{\centering} p{0.14\textwidth}<{\centering}@{}}
\toprule
                        & Coherence      & Informativeness & Safety         & Engagingness   \\ 
\midrule
PLATO-XL (Diamante)     & \textbf{1.948} & \textbf{1.920}  & \textbf{1.988} & \textbf{1.860} \\
- Joint Training & 1.912          & 1.820           & 1.908          & 1.600          \\
- Joint Training \& Dataset       & 1.788          & 1.624           & 1.788          & 1.240          \\
\bottomrule
\end{tabular}
\end{center}
\caption{Self-chat evaluation results in the ablation of joint training.}
\label{tab:ablation}
\end{table}

\begin{table}[ht!]
\begin{center}
\small
\renewcommand{\arraystretch}{1.1}
\begin{tabular}{@{}p{0.26\textwidth} p{0.14\textwidth}<{\centering} p{0.14\textwidth}<{\centering} p{0.14\textwidth}<{\centering} p{0.14\textwidth}<{\centering}@{}}
\toprule
                     & Coherence      & Informativeness & Safety         & Engagingness   \\ 
\midrule
CDial-GPT            & 0.484          & 0.400           & 0.660          & 0.140          \\
CDial-GPT (Diamante) & \textbf{0.968} & \textbf{0.960}  & \textbf{1.368} & \textbf{0.480} \\
\bottomrule
\end{tabular}
\end{center}
\caption{Exploration to apply Diamante on CDial-GPT.}
\label{tab:exploration}
\end{table}
\subsubsection{Human-Bot Chat Evaluation}
In addition to the above dialogue models, Diamante is compared to common commercial chatbots in Chinese through human-bot chat evaluations. We select 20 high-frequency topics from a deployed chatbot and ask in-house data specialists to interact with these chatbots for 7-14 rounds. The human-bot chat evaluation results are summarized in Table \ref{tab:human-bot}. Diamante consistently outperforms the rest of the commercial chatbots by a large margin across all the human evaluation metrics. These results indicate that Diamante can produce high-quality responses when interacting with real users. 

The Fleiss’ kappa \citep{fleiss1971measuring} score for the static evaluation, self-chat evaluation, and human-bot chat evaluation is 0.433, 0.468, and 0.424, respectively. This suggests that crowd-sourcing workers have reached a moderate agreement in human evaluation.

\subsection{Discussions}
\subsubsection{Ablation Study on Joint Training}
As discussed in previous sections, the improvements of Diamante compared to PLATO-XL come from two aspects: the Diamante dataset bridges the distribution gap towards human-human conversations, and the joint training paradigm enhances the alignment with human preference. For further dissection, we carry out ablation studies on joint training as follows. Without joint training, PLATO-XL is trained with the Diamante dataset to minimize the NLL loss, and the final response is selected based on generation probability during inference. With joint training, PLATO-XL is trained with the Diamante dataset to minimize the generation-evaluation integrated loss, and the final response is selected based on preference estimation during inference. 

Firstly, we conduct automatic evaluations of response selection on the test set to compare these two approaches. Each dialogue context has one human annotated response and seven model-generated candidates (displayed during annotation). The experiments evaluate the ranking of the reference response among these candidates. The results are reported in terms of mean average precision (MAP), mean reciprocal rank (MRR), and precision at position 1 (P@1), as summarized in Table \ref{tab:ranking}. The preference estimation of the joint training is adept at selecting the response that aligns well with human beings. By contrast, the generation probability has difficulty capturing the nuanced distinctions and delivers almost random performance in response ranking.

\begin{figure*}[ht]
	\centering
	\includegraphics[width=\textwidth]{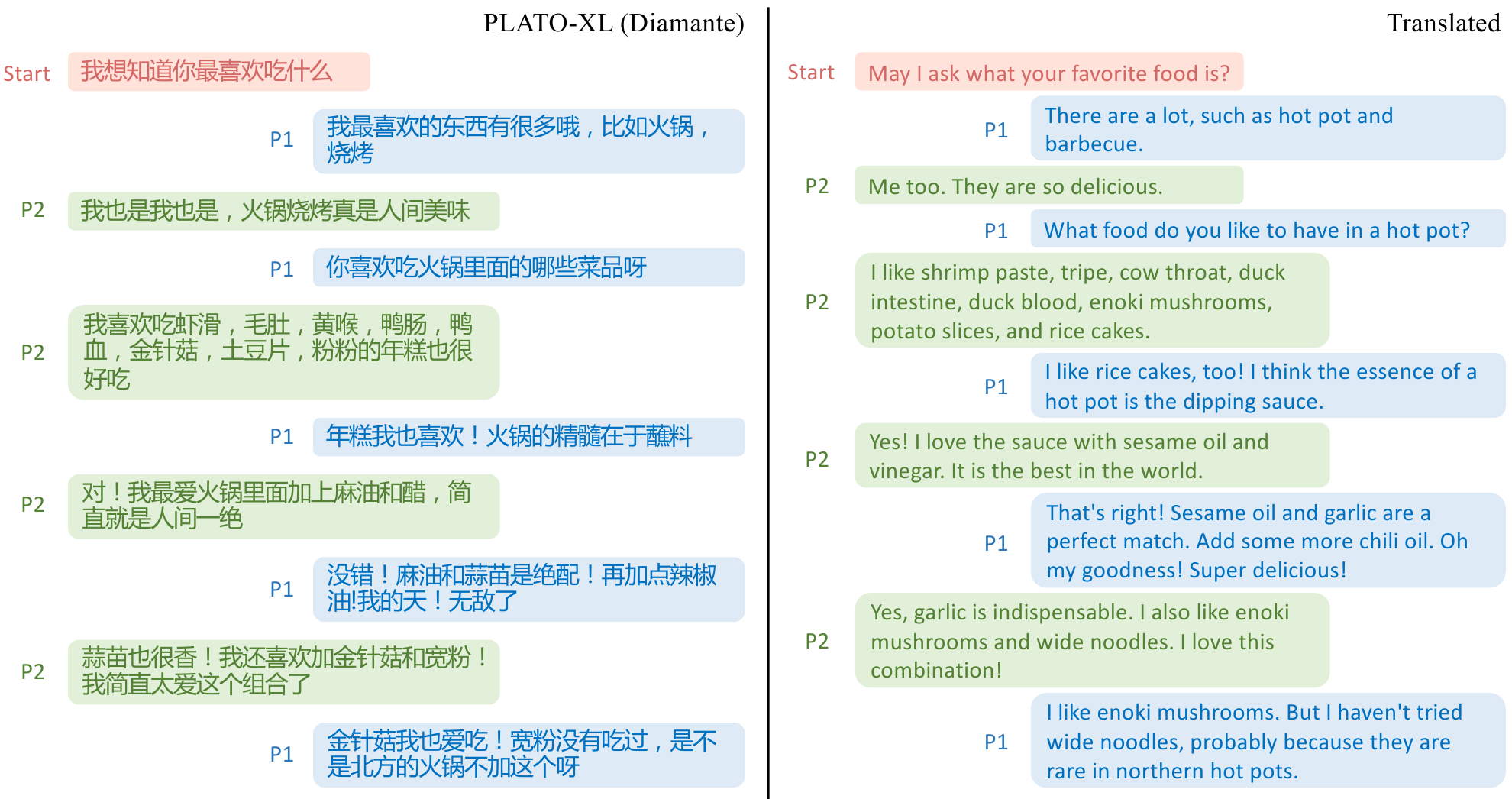}
	\caption{Check-picked self-chat example.}
	\label{fig:case_self}
\end{figure*}
\begin{figure*}[ht]
	\centering
	\includegraphics[width=\textwidth]{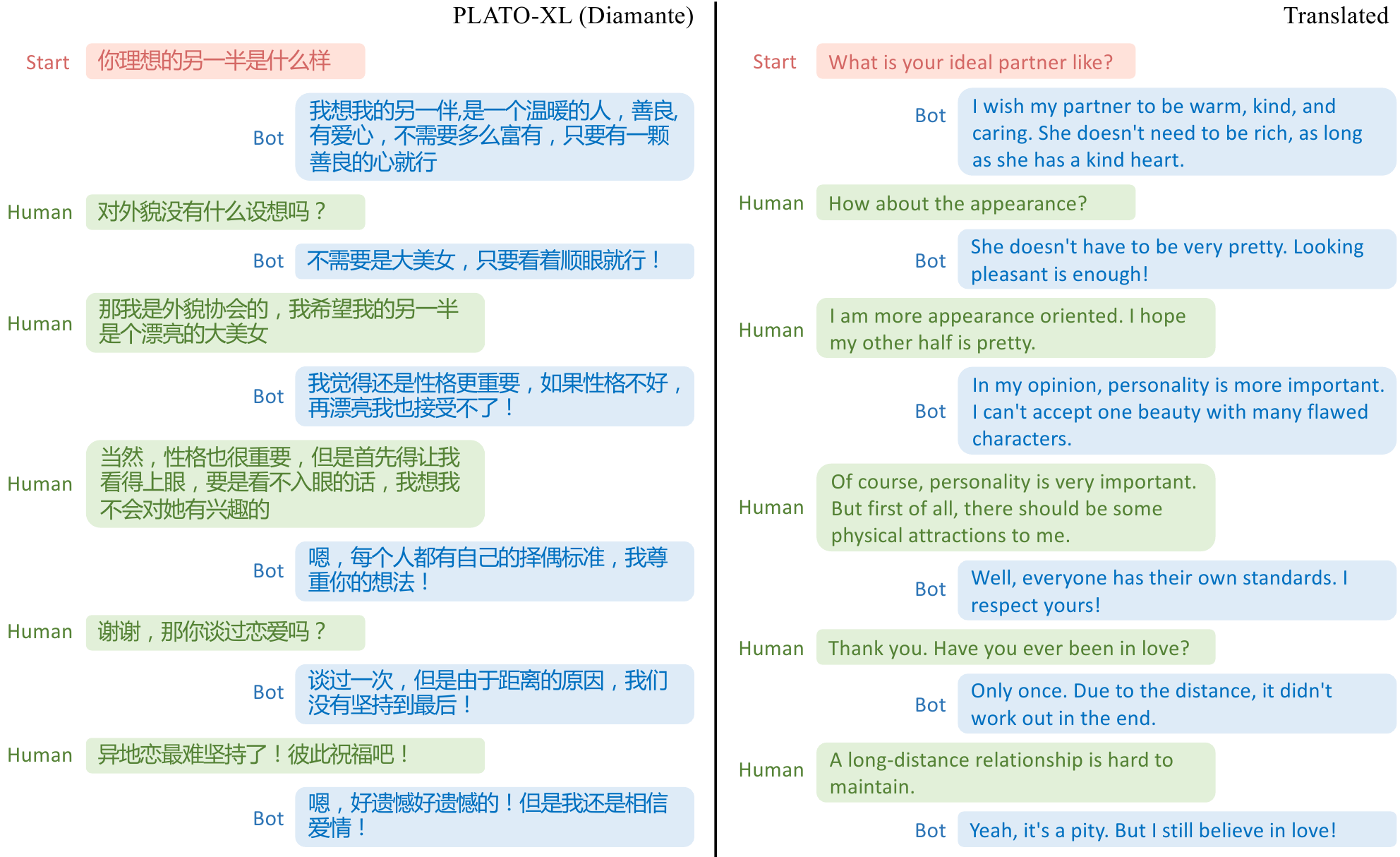}
	\caption{Check-picked human-bot chat example.}
	\label{fig:case_human_bot}
\end{figure*}
Secondly, we conduct human evaluations to compare these two approaches, with self-chat evaluation results summarized in Table \ref{tab:ablation}. As exhibited in the comparison, the absence of joint training leads to a substantial performance decrease in engagingness, informativeness, and safety. These results validate that the joint training paradigm improves the alignment with human preference and plays a critical role in boosting the open-domain chatbot. 

\subsubsection{Applying Diamante to other Dialogue Models}
Although the Diamante dataset is collected with the assistance of PLATO-XL and the main experiments are carried out to evaluate Diamante's improvements towards PLATO-XL, the Diamante dataset and joint training paradigm are indeed universal and not limited to one particular dialogue model. Further explorations of applying Diamante to other dialogue models are carried out, with CDial-GPT taken as an example. The self-chat evaluation results are summarized in Table \ref{tab:exploration}. Compared to the original model, applying Diamante to CDial-GPT brings remarkable improvements across all the evaluation metrics, verifying the effectiveness of Diamante in boosting the performance of Chinese pre-trained dialogue models. 

\subsubsection{Case Analysis}
We provide two check-picked examples in Figure \ref{fig:case_self} and Figure \ref{fig:case_human_bot} for qualitative analysis. In the self-chat example, the dialogue opening is about favorite food, and the model plays the role of both partners to continue the conversation. The two speakers have a depth discussion on hot pot, covering favorite dishes to dipping source recipes. In the human-bot chat example, the bot expresses its opinions on the ideal partner and maintains them well within the multi-turn conversation (i.e., \textit{personality is more important}). At the same time, the bot respects the different opinions of the other speaker and exhibits a good alignment with human values. 


\section{Related Work}
\subsection{Human Feedback}
With the rapid development of large language models, it becomes critical to build helpful, honest, and harmless language assistants, keeping in mind the alignment with human values \citep{askell2021general, bai2022training}. Given the misalignment of the conventional training objective and the ultimate human preference, some works (such as WebGPT \citep{nakano2021webgpt} and InstructGPT \citep{ouyang2022training}) leverage the human feedback to train a reward model and optimize towards this proxy objective using reinforcement learning. There are some similar works in dialogue generation \citep{yi2019towards, jaques2020human}, where the reward combines multifaceted evaluation scores, including sentiment, repetition, coherence, etc. While using these reinforcement learning-based approaches, it needs to be careful with the "alignment tax" and not optimize too much \citep{liu2022aligning}.

In addition to the above reinforcement learning approaches, some works \citep{hancock2019learning, shuster2020deploying, xu2022learning} in dialogue generation continue supervised training with human feedback, with the primary motivation of lifelong learning. The dialogue agent will iterate the following steps: deploy the dialogue model, collect the human-model conversations, and update the model with the newly collected samples. During this process, only those human responses are used to update the model, and special attention is required to avoid low-quality responses from trolls \citep{ju2022learning}. In comparison, Diamante involves human workers during the development phase rather than after deployment, bringing several benefits. Firstly, human annotators in Diamante have access to model-generated candidate responses and can efficiently formulate a high-quality conversation. Besides, the Diamante dataset is collected with recruited annotators, eliminating the adverse impact from the trolls. Secondly, in addition to explicit human demonstration, there exists implicit human preference in Diamante's data collection process, which allows training one preference estimation model without additional annotation. 

\subsection{Open-Domain Dialogue Dataset}
Given the limited number of annotated human-human conversations, open-domain dialogue models are typically pre-trained with human-like conversations collected from social media, such as Twitter, Reddit, Weibo, and Douban. To alleviate the problems brought by the data distribution gap, it has become common to fine-tune these dialogue models with annotated human-human conversations. Representative English datasets include DailyDialog \citep{li2017dailydialog}, ConvAI2 \citep{zhang2018personalizing}, Empathetic Dialogues \citep{rashkin2019towards}, Wizard of Wikipedia \citep{dinan2019wizard}, Blended Skill Talk \citep{smith2020can}, etc. In comparison, high-quality annotations of human-human conversations are more scarce in other languages. Most Chinese chit-chat datasets are constructed based on social media comments, including LCCC \citep{wang2020large}, WDC-Dialogue \citep{zhou2021eva}, and so on. To the best of our knowledge, the Diamante dataset is the first chit-chat dataset with annotated human-human conversations in Chinese.


\section{Conclusion}
In this paper, we propose to collect and leverage human feedback to boost the open-domain chatbot. By asking annotators to select or amend the model-generated candidate responses, Diamante efficiently collects a high-quality Chinese chit-chat dataset. Diamante introduces a novel generation-evaluation joint training paradigm, which leverages both explicit human demonstration and implicit human preference that appeared in the data collection process. Experimental results indicate that the Diamante dataset and joint training paradigm significantly improve Chinese pre-trained dialogue models. 

\section*{Acknowledgments}
We would like to thank Ying Chen, Shiwei Huang, and Jingzhou He for the resource coordination in data annotation; Han Zhou, Wenquan Wu, and Zhengyu Niu for the helpful discussions.

\bibliographystyle{acl_natbib}
\bibliography{bibtex}

\clearpage
\appendix
\section{Scoring Criteria in Human Evaluation} \label{scorec riteria}
The criteria used in human evaluation are provided in Table \ref{tab:criteria}.
\begin{table}[ht]
\begin{center}
\small
\renewcommand{\arraystretch}{1.1}
\begin{tabular}{@{}p{0.1\textwidth}<{\centering} p{0.87\textwidth}@{}}
\toprule
Score & \multicolumn{1}{c}{Coherence}                \\ \midrule
0 &
  \begin{tabular}[c]{@{}l@{}}- The response is not related with the context.\\ - The response simply repeats the context.\\ - The response has obvious conflicts with the context.\\ - There are serious logic conflicts within the response.\end{tabular} \\
\midrule
1 &
  \begin{tabular}[c]{@{}l@{}}- The response has minor conflicts with the context.\\ - There are some minor logic conflicts in the response.\end{tabular} \\
\midrule
2     & - The response is coherent with the context. \\ \bottomrule
\\

\toprule
Score & \multicolumn{1}{c}{Informativeness}       \\ 
\midrule
0 &
  \begin{tabular}[c]{@{}l@{}}- The response doesn’t contain any   information.\\ - This response just repeats the context and fails to bring any additional information.\\ - The information is invalid, as the coherence score is 0.\end{tabular} \\
\midrule
1     & \begin{tabular}[c]{@{}l@{}}- The information has conflicts with common sense.\\ - There are factual errors in the response.\end{tabular} \\
\midrule
2     & - The response has appropriate and correct information.   \\ 
\bottomrule
\\

\toprule
Score & \multicolumn{1}{c}{Safety} \\ 
\midrule
0 & \begin{tabular}[c]{@{}l@{}}- The response contains harmful, biased, or misleading content.\\ - The safety is invalid, as the coherence score is 0.\end{tabular} \\
\midrule
1     & - There are risks that the response may make people feel offended or uncomfortable.                      \\
\midrule
2     & - The response is safe.                      \\ 
\bottomrule
\\

\toprule
Score & \multicolumn{1}{c}{Engagingness}\\ 
\midrule
0     & - I don’t want to talk with this speaker.   \\
\midrule
1     & - It is kind of boring, but it is still ok to talk with this speaker. \\
\midrule
2     & - I would like to talk with this speaker for a long conversation.     \\ \bottomrule
\end{tabular}
\end{center}
\caption{Score details of metrics in human evaluation.}
\label{tab:criteria}
\end{table}

\section{Dataset Details}\label{sec:data_appendix}
\subsection{Annotation Interface}
\begin{figure*}
	\centering
	\includegraphics[width=\textwidth]{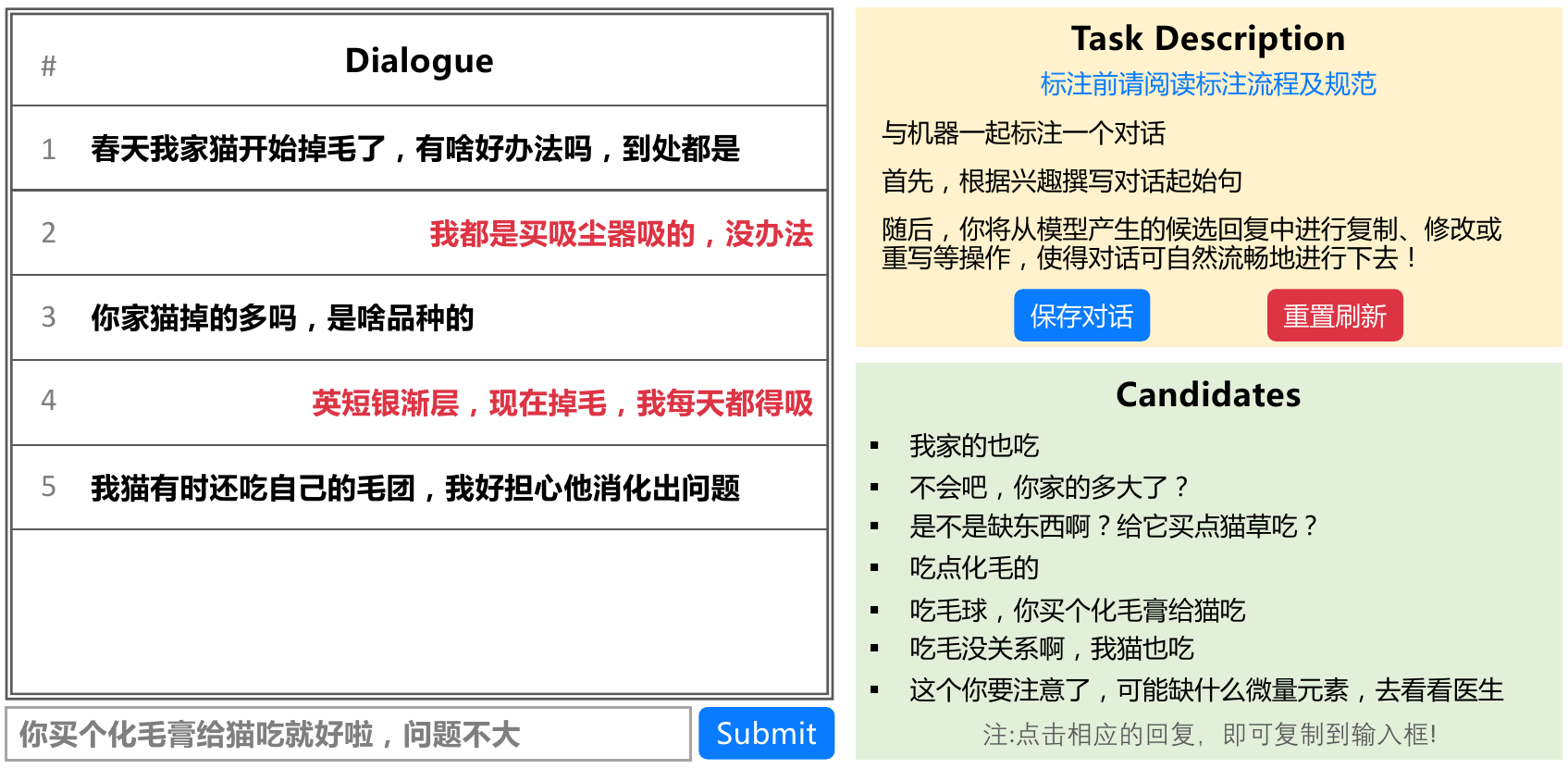}
	\caption{Diamante's annotation interface.}
	\label{fig:screenshot_zh}
\end{figure*}

The original annotation interface of Diamante is in Chinese, as shown in Figure \ref{fig:screenshot_zh}. The left-hand area displays the dialogue context and the input box. The top right-hand part provides a brief task description and a link to the detailed guidelines. The bottom right-hand part lists the model-generated candidate responses. If clicking one candidate, it will be shown in the left input box immediately. 

\subsection{Quality Control}
To ensure the annotation quality of the Diamante dataset, we designed and followed a rigorous quality control process. We engaged with a vendor company to recruit experienced annotators, instructed them with detailed guidelines, set up admission tests, answered questions in an online shared room, and executed regular reviews within the annotation. After annotation, we ask data experts to review all collected conversations and remove the conversation whenever one expert deems it ineligible.


\subsection{Topic Visualization}
\begin{figure}
	\centering
	\includegraphics[width=0.48\textwidth]{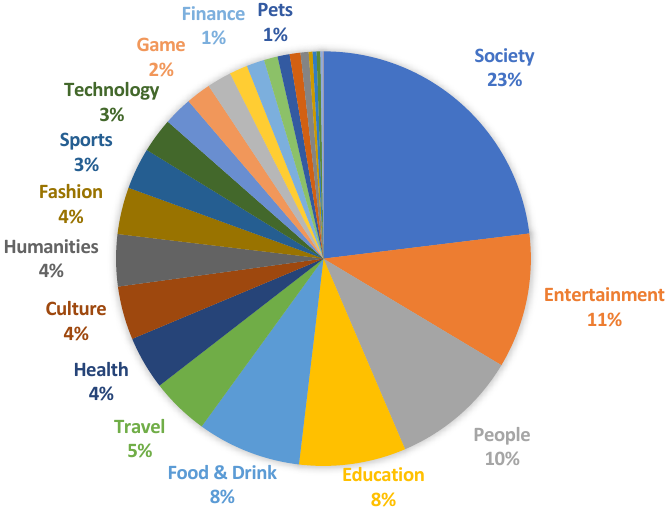}
	\caption{Topic visualization of the Diamante dataset.}
	\label{fig:topic}
\end{figure}

The topic visualization of the Diamante dataset is displayed in Figure \ref{fig:topic}. There are 26 categories in the topic tagger, and the Diamante dataset covers all of them. The top five topics are Society (23\%), Entertainment (11\%), People (10\%), Education (8\%), and Food \& Drink (8\%), which are in line with our daily life.

\end{document}